\documentclass[conference]{IEEEtran}
\IEEEoverridecommandlockouts
\usepackage{cite}
\usepackage{amsfonts}
\usepackage{algorithmicx}
\usepackage{graphicx}
\usepackage{textcomp}
\usepackage{xcolor}
\def\BibTeX{{\rm B\kern-.05em{\sc i\kern-.025em b}\kern-.08em
    T\kern-.1667em\lower.7ex\hbox{E}\kern-.125emX}}

\usepackage{booktabs} 
\usepackage{algorithm}
\usepackage{algpseudocode}
\usepackage{amsmath}
\usepackage{amssymb}
\usepackage{url}

\usepackage{array}

\usepackage{tikz}
\usetikzlibrary{calc}
\usepackage{algpseudocode}
\usepackage{comment}

\usepackage{braket} 
\usepackage{algorithm,algpseudocode}
\usepackage{blkarray}
\usepackage{multirow}

\usepackage{fancyhdr}
\usepackage{blkarray}
\usepackage{textcomp}
\usepackage{xcolor}

\author{\IEEEauthorblockN{1\textsuperscript{st} Given Name Surname: Anonymous}}

\begin{document}
\title{Hybrid Ensemble Deep Graph Temporal Clustering for Spatiotemporal Data
\thanks{This work was supported by iHARP: NSF HDR Institute for Harnessing Data and Model Revolution in the Polar Regions}
}

\author{
\IEEEauthorblockN{
  Francis Ndikum Nji$^1$,
  Omar Faruque$^1$,
  Mostafa Cham$^1$,
  Janeja Vandana$^1$,
  Jianwu Wang$^1$}
  
  $^1$Department of Information Systems,
      University of Maryland, Baltimore County,
      Baltimore, MD, United States\\
  Email: \{fnji1, omarf1, mcham2, vjaneja, jianwu\}@umbc.edu\\
  
}

\maketitle

\begin{abstract}
Classifying subsets based on spatial and temporal features is crucial to the analysis of spatiotemporal data given the inherent spatial and temporal variability. Since no single clustering algorithm ensures optimal results, researchers have increasingly explored the effectiveness of ensemble approaches. Ensemble clustering has attracted much attention due to increased diversity, better generalization and overall improved clustering performance. While ensemble clustering may yield promising results on simple datasets, it has not been fully explored on complex multivariate spatiotemporal data. For our contribution in this field, we propose a novel hybrid ensemble deep graph temporal clustering (HEDGTC) method for multivariate spatiotemporal data. HEDGTC integrates homogeneous and heterogeneous ensemble methods and adopts a dual consensus approach to address noise and misclassification from traditional clustering. It further applies graph attention autoencoder network to improve clustering performance and stability. When evaluated on three real-world multivariate spatiotemporal data, HEDGTC outperforms state-of-the-art ensemble clustering models by showing improved performance and stability with consistent results. This indicates that HEDGTC can effectively capture implicit temporal patterns in complex spatiotemporal data.
\end{abstract}


\section{Introduction}\label{sec:intro} Ensemble clustering is a burgeoning subfield in unsupervised machine learning that offers a powerful approach to address complex challenges by merging the results of multiple base clustering algorithms to generate a final clustering. Motivated by the success of ensemble approaches in supervised learning~\cite{kittler1998combining}, ensemble clustering was proposed to improve robustness and diversity through harnessing the collective intelligence of multiple clustering algorithms. However, finding a consensus from base clustering algorithms is difficult due to the following reasons; different number of clusters, cluster label ambiguity, varying shapes and sizes of clusters, overlapping clusters, diversity in clustering algorithms, aggregation complexity and the lack of ground truth data. The concept of ensemble clustering has been intensively investigated in applications, such as multimedia~\cite{yang2018adaptive,shi2018transfer}, pattern recognition~\cite{liu2018feature,tao2019marginalized}, and bioinformatics~\cite{yu2014adaptive,kiselev2017sc3}. Some execute a collection of different clustering algorithms (\textit{heterogeneous ensemble} ~\cite{yoon2006heterogeneous}), while others execute a single clustering algorithm multiple times with different initializations (\textit{homogeneous ensemble}~\cite{boongoen2018cluster}). 

Despite extensive research, ensemble temporal clustering remains underexplored in the context of complex multivariate spatiotemporal data, with no existing literature specifically addressing this application. Temporal clustering is crucial in many fields, such as traffic management, crop science and climate science, where understanding temporal patterns is vital. In climate science, take the study of climate variability in the Arctic as an example. Arctic sea ice, which significantly influences Earth's heat and freshwater balance, has shown a sharp long-term decline amidst strong internal variability. This variability affects atmospheric conditions, potentially impacting mid-latitude weather patterns often resulting to extreme events. One way to understand the reasons for this sharp longterm decline is identifying temporal patterns through cluster analysis in Arctic sea ice variability. Current clustering approaches are mostly distance-based and face significant challenges due to inability to properly handle noise and outliers, high dimensionality and complexity of data resulting from the non-straightforward, intricate interplay between spatial and temporal dimensions. Ensemble approaches were introduced to mitigate some of these challenges and improve clustering results. Unfortunately they continue to face significant challenges due to their 
inability to capture non-linear relationships from complex physical interactions, failure to capture the temporal dependencies and trends. To address these challenges, we propose a Hybrid Ensemble Deep Graph Temporal Clustering Framework designed to categorize multivariate spatiotemporal data into meaningful. Our proposed model leverages the strengths of both homogeneous and heterogeneous ensemble approaches. In homogeneous ensemble clustering, multiple clustering models of the same type are used to create an ensemble with the goal of improving robustness, stability, and accuracy. On the other hand, Heterogeneous ensemble clustering uses multiple clustering models of different types to produce clustering with a goal of improving diversity and performance. In the same sense, we leverage the strength of both the co-occurrence consensus and non-negative matrix factorization consensus to further enhance model performance by reducing noise and misclassification errors. The resulting matrix is fed into a graph attention autoencoder which performs KMeans clustering on the latent space to obtain our final partitions. Our contributions to address existing ensemble clustering challenges on high dimensional spatiotemporal data are summarized as follows.
\begin{itemize}
    \item To the best of our knowledge, we are the first to propose an end-to-end ensemble clustering framework for complex multivariate spatiotemporal data that integrates homogeneous and heterogeneous ensemble techniques to mitigate existing drawbacks with improved performance and stability of final clusters.
    \item We introduce a meta consensus layer in our ensemble model that integrates approaches from the co-occurrence consensus category and median partition consensus category to reduce noise and misclassified samples.
    \item To capture temporal neighborhood and temporal dynamics, we introduce a stacked graph attention network equipped with three GATv2 and two LSTM layers.
    \item Extensive experiments conducted on three real-world multivariate spatiotemporal datasets demonstrate improved performance and stability of our proposed model over state-of-the-art ensemble clustering models. Our implementation code is publicly available\footnote{https://github.com/big-data-lab-umbc/multivariate-weather-data-clustering/tree/main/HESC}.
    
\end{itemize}




\section{Related Work}\label{sec:Related}
\textbf{Heterogeneous Ensemble Clustering.}
This approach seeks to eliminate the drawbacks of single clustering solutions by consolidating the results from multiple \textit{diverse} base clustering algorithms.
Pfeifer et al.~\cite{PFEIFER2023104406} proposed \(Parea_{hc}\), a multi-view hierarchical heterogeneous ensemble clustering approach for disease subtype detection with hierarchical agglomerative clustering and spectral clustering.
While \(Parea_{hc}\) promises improved accuracy, it still suffers scalability issues when applied to very large datasets. Furthermore, since it is limited to only two fixed algorithms, the final results may suffer from algorithm selection bias which can greatly limit its robustness, stability, performance and generalizability.
Recently, Miklautz et al.~\cite{miklautz2022deep} proposed Deep Clustering With Consensus Representations (DECCS) that repeatedly learns a consensus representation and subsequently a consensus clustering. While DECCS may improve clustering, it greatly relies on consensus representations and hence may face limitations when dealing with data whose sample labels are previously not known. Moreover, it relies on certain assumptions about the data, and could produce sub-optimal results for large-scale complex spatiotemporal data.
Bedali et al.~\cite{bedalli2016heterogeneous} proposed a heterogeneous cluster ensemble approach to improve the stability of fuzzy cluster analysis. Their approach uses four fuzzy clustering models whose results are later merged through a consensus matrix to obtain the final partitions. Although they used four fuzzy algorithms for ensemble, their approach did not account for noise and misclassification from these algorithms.

\textbf{Homogeneous Ensemble Clustering.}
Caruana et al.~\cite{caruana2006meta} proposed meta clustering which creates a new mode of interaction between users, the clustering system, and the data. It uses KMeans to generate multiple base partitions which are themselves clustered at the meta level following the co-occurrence approach. Meta clustering suffers from the initial seeds problem, cluster structure preservation, spherical clusters and categorical data. Liu et al.~\cite{liu2015spectral} proposed the spectral ensemble clustering (SEC) which attempts to mitigate the relative high time and space complexity when applying the co-occurrence matrix on a dataset.
While SEC may benefit from reduced space and time complexity, it struggles to effectively cluster multi-scale data with different cluster sizes and densities. 
\textbf{Graph Attention Network}. Graph attention networks (GATs) leverage masked self-attentional layers to address the shortcomings of prior methods which are based on graph convolutions or their approximations. Without prior knowledge of the input graph structure and without requiring any kind of costly matrix operation, GAT's are able to attend over their neighborhoods’ features by specifying different weights to different nodes in a neighborhood. Recently, GAT has gained much popularity with its potential well demonstrated in various areas. Wang et al., \cite{wang2019heterogeneous} recently proposed a heterogeneous graph neural network (HAN) based on the hierarchical attention, including node-level and semantic-level attentions. The node-level attention aims to learn the importance between a node and its meta-path based neighbors, while the semantic-level attention is able to learn the importance of different meta-paths. With the learned importance from both node-level and semantic-level attention, the importance of node and meta-path can be fully considered. Xie et al., \cite{xie2020mgat} proposed a Multi-view Graph Attention Networks (MGAT). They explore an attention-based architecture for learning node representations from each single view, the network parameters of which are constrained by a novel regularization term. To collaboratively integrate multiple types of relationships in different views, a view-focused attention method is explored to aggregate the view-wise node representations. More recently, Brody et al., proposed GATv2: a dynamic graph attention variant of GAT, which addresses GATs static attention problem by modifying the order of operation. Salehi et al., \cite{salehi2019graph} proposed graph attention auto-encoder (GATE), capable of reconstructing graph structured inputs including both node attributes and the graph structure, through stacked encoder/decoder layers equipped with self-attention mechanisms. By considering node attributes as initial node representations in the encoder, each layer generates new representations of nodes by attending over their neighbors’ representations. This process is reversed to reconstruct node attributes.\\

\section{Problem Definition}\label{sec:prob}
Given unlabeled multivariate spatio-temporal climate data, and without prior knowledge of sub-group memberships, our goal is to efficiently partition the data into distinct sub-groups based on temporal similarities among data points. To be specific, assume $n$ atmospheric variables ($x_i$) measured over a grid region covering $L$ longitudes and $W$ latitudes and stored in a vector $X=\{x_1, x_2, x_3, ..., x_n\}$ such that, for each time step every grid location has $n$ values for all variables. Variables are measured for $T$ different time steps, $X_i=\{x_1, x_2, x_3,..., x_n\}$, $i\in \{1, ...,T\}$.\\

\textbf{Input:} \( Dataset = \{X_1, X_2, X_3, ..., X_T\}\),
\begin{equation}
\begin{split}
X_{i} & =\begin{Bmatrix}
\begin{bmatrix}
 x_{1}(1,1)&x_{1}(1,2)  & \cdots  & x_{1}(1,W) \\ 
 x_{1}(2,1)&x_{1}(2,2)  & \cdots  & x_{1}(2,W) \\ 
 \vdots &  \vdots  & \ddots  &\vdots  \\ 
 x_{1}(L,1)&x_{1}(L,2)  & \cdots  & x_{1}(L,W)
\end{bmatrix} \vspace{0.2cm}\\ 
\begin{bmatrix}
 x_{2}(1,1)&x_{2}(1,2)  & \cdots  & x_{2}(1,W) \\ 
 x_{2}(2,1)&x_{2}(2,2)  & \cdots  & x_{2}(2,W) \\ 
 \vdots &  \vdots  & \ddots  &\vdots  \\ 
 x_{2}(L,1)&x_{2}(L,2)  & \cdots  & x_{2}(L,W)
\end{bmatrix} \\
 \vdots \\
\begin{bmatrix}
 x_{n}(1,1)&x_{n}(1,2)  & \cdots  & x_{n}(1,W) \\ 
 x_{n}(2,1)&x_{n}(2,2)  & \cdots  & x_{n}(2,W) \\ 
 \vdots &  \vdots  & \ddots  &\vdots  \\ 
 x_{n}(L,1)&x_{n}(L,2)  & \cdots  & x_{n}(L,W)
\end{bmatrix} 
\end{Bmatrix}
\end{split}
\end{equation}
\noindent , where \(X_i\) represents one observation, \(x\) represents one variable of an observation, \(i \in \{1, ..., T\}\), \(T\) represents the number of time steps, \(n\) represents the number of atmospheric variables, \(L\) and \(W\) represent the longitude and latitude respectively.\\

\textbf{Output:} Our proposed clustering model should partition the \(Dataset = \{X_1, X_2,..., X_T\}\) into \(k\) clusters: \(C_1, C_2, \dots, C_k\), where\(k<T\), such that objects within the same cluster are similar to each other and dissimilar to those in other clusters.\\ Formally:\begin{equation*}
\begin{split}
    C_1=\{X_{C_1}^1, X_{C_1}^2,& ..., X_{C_1}^{n_1}\}, C_2=\{X_{C_2}^1, X_{C_2}^2, ..., X_{C_2}^{n_2}\}, ..., \\
    &C_k=\{X_{C_k}^1, X_{C_k}^2, ..., X_{C_k}^{n_k}\} 
\end{split}    
\end{equation*}
\begin{equation*}
X_{C_j}^{i} \in X, i\in \{1, ..., n_j\}, j\in \{1, ..., k\} 
\end{equation*} 
\noindent , where \(n_j\) = number of observations of cluster \(j\).
\begin{equation*}
\bigcup_{j=1}^{k}C_j=X \textrm{ and } C_j\cap C_l=\varnothing
\end{equation*}
Here, \(j\neq l\) and \((j, l)\in \{1, ..., k\}\) for each pairs of clusters.

\section{Proposed Framework}\label{sec:Framework}
\begin{figure*}[ht]
  \centering
  \includegraphics[width=0.8\textwidth]{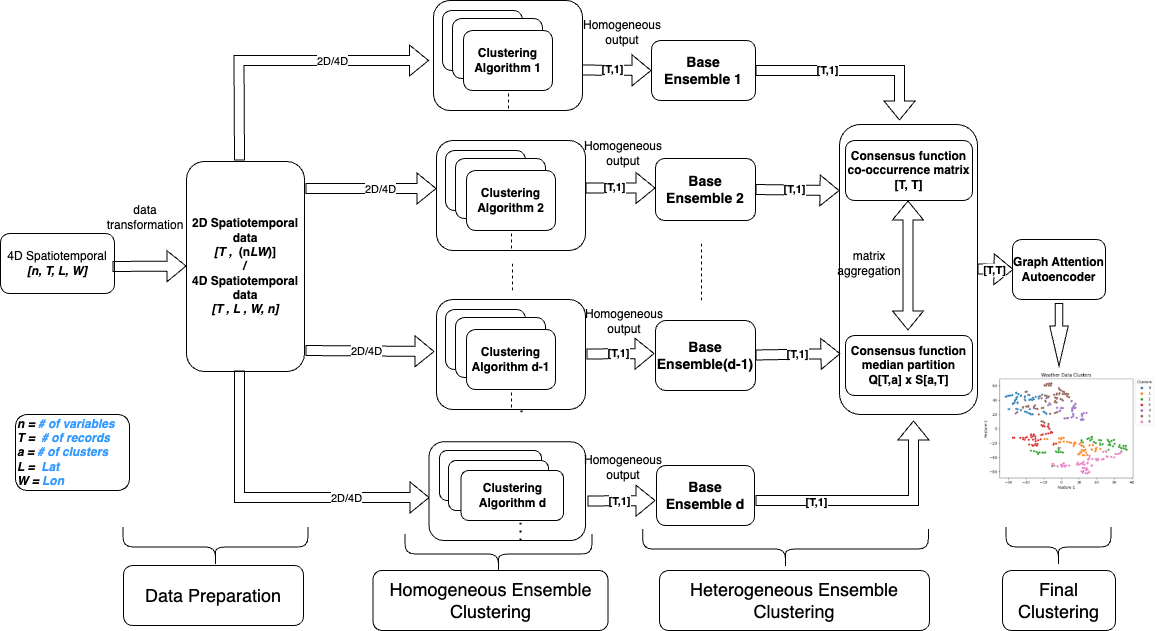}
  \centering \caption{\textbf{Architecture of our proposed Hybrid Ensemble Deep Graph Temporal Clustering (HEDGTC) model}; This is an end-to-end architectural flow diagram representing various phases of our proposed model. The process starts with the data preparation phase were we data is injected and preprocessed. Homogeneous Ensemble Clustering represents the second phase and individually executes all base clustering models in a homogeneous fashion. Heterogeneous Ensemble Clustering consolidates the clustering results from the previous phase through co-occurrence consensus and the non-negative matrix factorization. Further merging of the resulting matrices is done to yield one combined matrix. The Final Clustering phase applies a graph attention auto encoder on the combined matrix providing our final partitions.}
  \label{fig:arch}
\end{figure*}

In this paper, we propose a novel Hybrid Ensemble Deep Graph Temporal Clustering (HEDGTC) model capable of clustering unlabeled multivariate spatiotemporal datasets.
Our goal is to generate highly distinct and different member clusters from across base partitions. A high level of accuracy and diversity among resulting clusters imply they have successfully captured distinct temporal information about the data and can potentially improve the performance and stability of our model. The performance of base learners is a huge determinant for the performance and accuracy of our final clustering~\cite{caruana2006meta}. Our model selection approach is rigorous and selects best \(k\) performing conventional and deep clustering clustering algorithms across a different categories of clustering algorithms, were \(k\) is an integer and represents the number of base clustering algorithms. Figure~\ref{fig:arch} presents an overview of our proposed framework. HEDGTC is made up of four interconnected phases: \textit{data preparation, homogeneous ensemble clustering, heterogeneous ensemble clustering and final clustering}.

\subsection{Data Preparation}\label{sec:data_prep} This is the first phase of our model pipeline and involves the collection and preparation of data needed for our experiment. We limit our scope to real-world data collected over space and time through weather sensors and satellites, and stored in systems like NOAA Physical Sciences Laboratory (PSL) \cite{NCAR}, Meteorological Assimilation Data Ingest System (MADIS) \cite{MADIS} and Climate Data Store (CDS)\cite{ERA5}. This data is often in four dimensions (4D): time, longitude, latitude, and measured variables such as snowmelt, sea ice extent, and total cloud cover. Traditional distance-based clustering and evaluation algorithms usually have a hard time to accept 4D datasets as inputs. For instance, KMeans computes cluster centers based on the smallest squared Euclidean distances among data points and classifies these points to belong to a cluster. In the same sense, it is a challenge for distance-based clustering evaluation approaches such as silhouette score \cite{rousseeuw1987silhouettes} to accept 4-dimensional datasets.
Research in the area of directly applying distance-based clustering and evaluation methods on 4D data is still at its preliminary stage hence there exist few well tested algorithms. For this reason, we transform our raw 4D data structure into a 2D structure acceptable by our selected distance-based base clustering algorithms. 
For generalizability, as shown in Figure~\ref{fig:arch} our model transforms the raw data into both 4D [\(n, T, L, W\)] and 2D [\(T, (n,\ L ,\ W )\)] to be able to accommodate the input demands of various algorithms respectively. Further data processing is done by data rescaling, dealing with null and extreme values as explained in detail in Section \ref{sec:Experiments}.

\subsection{Homogeneous Ensemble Clustering}\label{model_select}
This is the second phase of our model pipeline. In Figure~\ref{fig:arch}, \textit{clustering algorithm d} represents each selected base clustering algorithm. HEDGTC is extensible and can accommodate an infinite number of base clustering models. Each selected algorithm at this phase performs homogeneous ensemble clustering whereby it is executed \(q\) times with different initialization and parameter settings to capture different aspects of the data, enhance robustness and leverage diversity among partitions\cite{ gionis2007clustering}. Here \(q\) is user specified. Two challenges at this phase are selecting the best clustering models for the ensemble; since different clustering models perform differently provided different datasets and determining the optimal number of clusters; since different algorithms compute clustering differently and different datasets contains different optimal number of clusters. To address these challenges, Distortion Score Elbow approach \cite{jain1988algorithms} is used to determine the optimal number of clusters. Base clustering algorithms are selected based on their performance and executed multiple times in a homogeneous manner with different initialization and hyperparameter settings. This is seen to improve stability. Consensus is done through the co-occurrence function to determine the final labels.
\subsection{Heterogeneous Ensemble Clustering}
A cluster ensemble is heterogeneous if the partitions in the previous stage are obtained from applying different clustering algorithms\cite{ayad2011heterogeneous, hou2013heterogeneous}. From Figure \ref{fig:arch}, \textit{Base Ensemble 1 to d} represent heterogeneous clustering results as 1D [\(T, 1\)] vectors. These results are consolidated through voting onto a [\(T,  T\)] matrix, where \(T\) is the time step and size of the matrix. From literature, combining results from different clustering algorithms produces better clustering solutions with improved accuracy, performance and diversity~\cite{yoon2006heterogeneous}. Since we assume that a cluster structure exists in our dataset (clusterability~\cite{adolfsson2019cluster}), the goal of consolidation is to leverage the complementary strengths and mitigate the limitations of individual algorithms, capture different perspectives of the data, increase robustness and enhance the overall clustering performance. To fully maintain the existing cluster structure of the data, we merge the implementation of two distinct consensus approaches while harnessing their full benefits. Diversity of the base partitions is of great importance and is captured by specifying how different views of the cluster structure are disclosed by different base partitions. If each base partition discloses the cluster structure from different view-points, their consensus may indicate the global view of the structure. Hadjitodorov et al.~\cite{hadjitodorov2006moderate} showed that even moderate diversity of the base partitions could be more effective for cluster ensemble. To achieve an optimized heterogeneous ensemble, we merge clustering results from both object co-occurrence-based and median partition-based approaches. From the list of algorithms that follow object co-occurrence based approaches, we use the \textit{co-occurrence matrix} and from those that follow the median partition-based approaches, we use the \textit{non-negative matrix factorization based consensus} algorithm. These algorithms were chosen based on diversity and accuracy \cite{boongoen2018cluster, yang2016temporal}.

\textbf{Consensus through Object Co-occurrence.}\label{co-ass} Consensus clustering seeks to combine results from several clustering algorithms to increase the robustness of clustering analyses\cite{zhou2021clustering, shi2022consensus, liu2016infinite}. Object Co-occurrence is one of the two widely used approaches to reach consensus. From Figure \ref{fig:arch}, the clustering results from \textit{Base Ensemble 1 to d} each with dimensions [\(T, 1\)] are consolidated in a \textit{the co-association matrix} of dimension [\(T,  T\)] through the co-occurrence consensus function. This step was added to avoid the \textit{label correspondence problem} by mapping the ensemble members onto a new representation: \textit{the co-association matrix} in which a similarity matrix can be calculated between a pair of objects in terms of how many times a particular pair is clustered together in the base clustering.
\par Let \(P = \{P_1, P_2, \dots, P_M\}\) denote a set of \(M\) base partitions, where \(P_{i} = \{ C_{i1}, C_{i2}, \dots, C_{iK_{i}}\}\), and \(C_{ij}\) is the \(j^{th}\) cluster of \(p_i\), \(K_i\) is the number of clusters in \(P_i\). Suppose \(P = \{ C_1, C_2, \dots, C_K\}\) is the final clustering and \(K\) is the number of clusters in \(P\), \(C_i\) is a cluster of \(P\). Following~\cite{fred2005combining}, the co-association matrix is defined as:
\begin{equation}
    CM_{i,j} = \frac{1}{m} \sum_{m=1}^{M} \, \sum_{l=1}^{K_m} \mathcal{T}(i,j,C_{ml}) 
\end{equation}
\noindent , where \(CM_{i,j}\) denotes an entry of \(CM\), \(C_{ml}\) is the \(l^{th}\) base cluster
in \(P_m\), and \(\mathcal{T}(i,j,C_{ml})\) is an indicator:\\ \(\mathcal{T}(i,j,C_{ml}) = \begin{cases}
            1, & if \, \, \textbf{x}_i \in C_{ml} \wedge \textbf{x}_j \in C_{ml}\\
      0, & \text{otherwise}
         \end{cases}\)




The value in each position (\(i,j\)) of this matrix is a measure of how many times the objects \(x_i\) and \(x_j\) are in the same cluster for all partitions in \(\mathbb{P}\). 
To reduce noise and misclassification errors we apply matrix post-processing through matrix normalization by diagonalizing~\cite{kadison1984diagonalizing}, and applying a user-defined minimum threshold which reduces all connections or values less than our minimum set threshold to zero. 

\textbf{Consensus through Non-negative Matrix Factorization (NMF).}
The primary objective in clustering is given by: 
\begin{equation}
    \underset{\mathbb{C}_i}{\min}\sum_{i=0}^{k}\sum_{x \in S_{i}} ||(x_{i} - C_{i}||^2
\end{equation}
, where \(k\) is the number of clusters, \(S_{i}\) is the set of all points belonging to cluster \(i\), \(x\) is the data point and \(C_i\) is the \(i^{th}\) cluster. \((x - C_i)^2\) is the distance between the point \(x\) and the centroid \(C_i\).
To transform the clustering results from the objective function into a matrix, we define a new matrix named \(M\) of dimension \(k \times n\), where \(k\) is the number of centroids/clusters and \(n\) is the total number of data points: 

\begin{equation}\label{mat:matrix}
Z = \begin{blockarray}{cccccc}
    & x_1 & x_2 & x_j & \dots & x_n \\
\begin{block}{c(ccccc)}
C_1 &  &  & \downarrow  & \dots &  \\
C_2 &  &  & \downarrow   & \dots & \\
C_i & \rightarrow & \rightarrow & Z_{ij} & \dots &  \\
\vdots & \vdots & \vdots & \vdots & \ddots & \vdots \\
C_k &  &  &  & \dots &  \\     
\end{block}
\end{blockarray}
\substack{k \times n}
\end{equation}

Non-negative matrix factorization (NMF) \cite{cichocki2008advances} refers to the problem of factorizing a given non-negative data matrix \(M\) into two matrix factors \(A \) and \(B\), i.e., \(M \approx AB\), while requiring \(A\) and \(B\) to be non-negative \cite{li2007solving}. We use the following measure to compute distance between partitions: 





\begin{equation}\label{mp}
    \mu (P,P^\prime) = \sum_{i,j=1}^{n} \mu_{ij} (P,P^\prime)
\end{equation}
, where \(\mu_{ij} (P,P^\prime) = 1\) if \(x_i\) and \(x_j\) belong to the same cluster in partition \(P\) and belong to different clusters in partition \(P^\prime\), otherwise \(\mu_{ij} (P,P^\prime) = 0\).
Similarly, the connectivity matrix is expressed as:
\begin{equation}\label{conn}
    M_{ij}(P_v) = \begin{cases}
            1, & \exists C^v_t \in P_v \; s.t \; x_i \in C^v_t\, and\, x_j \in C^v_t;\\
      0, & \text{otherwise}
         \end{cases}
\end{equation}


Finally, the consensus clustering becomes the optimization problem: \(\underset{Q \geq 0, \, S \geq 0}{\min} \mid\mid \tilde{M} - QSP*T \mid\mid^2, \,s.t.\, \, Q^TQ = I\), where the matrix solution \(M\) is expressed in terms of the two matrices \(Q\) and \(S\), i.e, \(M \approx QS\). The optimization problem can then be solved using the following \textit{Multiplicative Update} rule (which is based on gradient descent with different update rules) \cite{lee2000algorithms, wang2012nonnegative, gillis2014and}:
\resizebox{\columnwidth}{!}{\(
\begin{aligned}
    Q_{ar} &\leftarrow Q_{ar} \sqrt{\frac{(\tilde{M}QS)_{ar}}{(QQ^T\tilde{M}QS)_{ar}}} \quad 
    S_{rb} &\leftarrow S_{rb} \sqrt{\frac{(Q^T\tilde{M}Q)_{rb}}{(Q^TQSQ^TQ)_{rb}}}
\end{aligned}
\)}

The \textit{Multiplicative Update Rule} is an iterative method and very sensitive to the initialization of \(Q\) and \(S\). In this paper, we use a Non-negative Double Singular Value Decomposition (NNDSVD) based initialization~\cite{boutsidis2008svd} to obtain \(Q\) and \(S\) and with these two matrices \(U = QSQ^T\) is obtained which is the connectivity matrix of the consensus partition \(P^*\). An important hyperparameter predefined is the rank denoted as \(r\) and determines the number of desired clusters as well as a condition for matrix multiplication for reconstructing the factorized matrices \(Q\) and \(S\). 

\textbf{Matrix Concatenation through Padding.} The output of our co-occurrence consensus approach is a 2-D [\(a,  a\)] square matrix were \(a = T\) while the output of our NMF consensus approach is a 2-D rectangular [\(a,  r\)] matrix were \(r\) represents the rank of decomposition. 

\begin{figure}[H]
  \centering
  \includegraphics[width=0.45\textwidth]{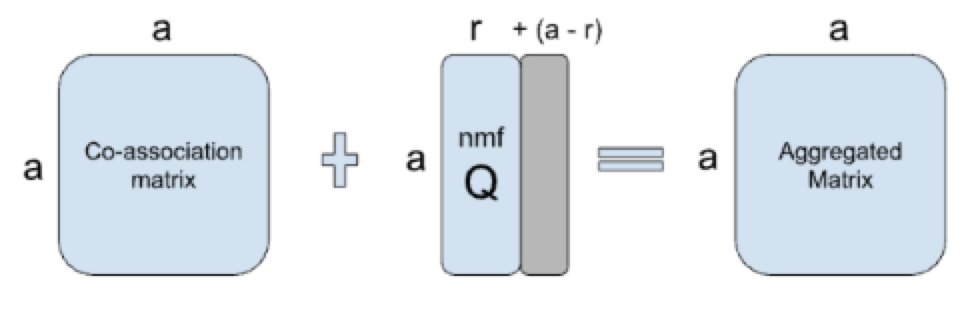}
  \centering \caption{Matrix Concatenation: The co-association matrix has dimension [\(a \times a\)] were \(a\) is the length of the time series, and the non-negative factorized matrix has dimension [\(a \times r\)] were \(r\) is the rank. \(Q\) is padded and added to the co=association matrix and the resulting matrix is of dimension [\(a \times a\)].}
  \label{fig:aggreg}
\end{figure}


\subsection{Graph Attention AutoEncoder based Final Clustering} In this section, we implemented the graph attention autoencoder to obtain our final partitions. The autoencoder is made up of stacked layers of GATv2 and LSTM layers for graph feature learning. Graph Attention Networks (GATs) leverage attention mechanisms for feature learning on graphs. Introduced by Veličković et al. \cite{velickovic2017graph}, GATs offer a more nuanced approach to aggregating neighborhood information. 

Recently, Brody et al. \cite{brody2021attentive} proposed GATv2, a variant of GAT that allows attention by changing the attention mechanism,
\begin{equation*}
    e_{ij} = \mathbf{a}^\top \text{LeakyReLU}\left(\mathbf{W} \left[\mathbf{h}_i \parallel \mathbf{h}_j\right]\right)
\end{equation*}
\begin{equation*}
    = \mathbf{a}^\top \text{LeakyReLU}\left(\mathbf{W}_l \mathbf{h}_i + \mathbf{W}_r \mathbf{h}_j\right)
\end{equation*}

\begin{figure}[ht!]
  \centering
  \includegraphics[width=0.49\textwidth]{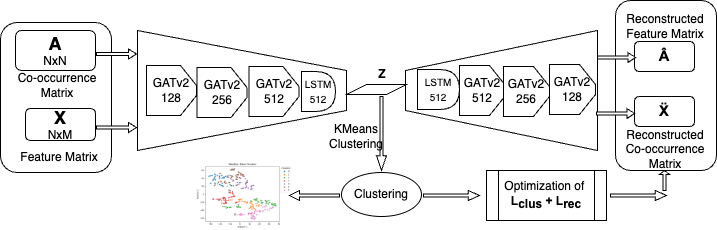}
  \centering \caption{Graph Attention AutoEncoder for clustering the final merged matrix: The input of the encoder is an adjacency matrix \(A\) and node features \(X\) and the output is the reconstructed \(\hat{\mathbf{A}}\) and \(\hat{\mathbf{X}}\). Input data is projected to a lower dimensional dense layer \(Z\) through stacked GATv2 and LSTM layers and KMeans is applied to the extracted features to generate our final clusterings.}
  \label{fig:gatAE}
\end{figure}

Figure \ref{fig:gatAE} depicts our proposed autoencoder architecture used for clustering the resulting merged matrix from the third phase of HEDGTC. The encoder takes the merged matrix \(A\) and feature matrix \(X\) as inputs and generates the latent variable \(Z\) as output. \(f_{\text{enc}}(\mathbf{A}, \mathbf{Z}) = \mathbf{Z}\). 

The encoder is composed of three stacked GATv2Conv layers, a TopKPooling layer, and two LSTM layers. Each GATv2Conv layer aggregates information from neighboring nodes using attention mechanisms, and refines attention coefficients iteratively, enabling the network to focus on different aspects of the graph structure. Stacking three layers allows the encoder to capture complex, multi-hop relationships between nodes, leading to richer feature representations and the ability to learn more abstract, high-level features. This setup also enables the network to capture long-range dependencies, which are crucial for tasks like clustering, where relationships between distant nodes are significant. Through testing, we found that using three layers strikes an optimal balance between model complexity, learning capacity, and computational efficiency. The TopKPooling layer is applied to rank and select the top \(k\) most important nodes, reducing dimensionality and enhancing the interpretability of the network's decisions. To fully capture temporal dependencies, we use two LSTM layers to cast the latent representation onto a more compact space in the temporal dimension. Finally, the latent features 
\(Z\) produced by the encoder are used with KMeans to assign data points into
\(k\) clusters, each with distinct temporal characteristics.

The decoder performs a reverse operation in a nonlinear manner by mapping the compact latent features from the encoder function into a new feature space that is identical to the input dataset \(\hat{A} \approx A\). The decoder process is constructed using stacked GATv2 and LSTM upsampling methods to effectively learn reconstruction parameters and minimize the difference between the reconstructed and input data: \(f_{\text{dec}}(\mathbf{Z}) = \hat{\mathbf{X}}, \hat{\mathbf{A}}\). Clustering is achieved by simultaneously learning a set of clusters in the latent feature space through the joint minimization of two objective functions. The first objective focuses on generating well-separated groups in the latent space by minimizing the clustering loss. The second objective aims to reduce the mean square error between the reconstructed data and the input dataset, known as reconstruction loss. By optimizing both objectives together, the autoencoder is guided to extract efficient temporal features that are well-suited for categorizing the input data into 
\(k\) clusters. The model loss is then computerd as follows: \(L = L_{\text{rec}}(\mathbf{X}, \hat{\mathbf{X}}, \mathbf{A}, \hat{\mathbf{A}}) + \lambda \, L_{\text{clus}}(\mathbf{Z})\), where \(L_{\text{rec}}\) is the reconstruction loss, which measures how well the decoder can reconstruct the input graph, \(L_{\text{clus}}\) is the clustering loss, which encourages the latent representation to form meaningful clusters and \(\lambda\) is a hyperparameter that controls the trade-off between reconstruction and clustering.

\section{MODEL EVALUATION METRICS}\label{sec:Evaluation}


\subsection{Performance Related Evaluation Metrics}
In the absence of ground truth, we evaluate the performance of our proposed model on six internal cluster validation measures. These measures seek to balance the \textit{compactness} and the \textit{separation} of formed clusters through minimizing intra-cluster distance and maximizing the inter-cluster distance respectively. They are:

\subsubsection*{Silhouette Score}
This index measures the normalised difference between the intracluster and intercluster average distances \cite{ rousseeuw1987silhouettes}. 
\subsubsection*{Davies-Bouldin score (DB)} This measures the average similarity measure of each cluster with its most similar cluster, where similarity is the ratio of within-cluster distances to between-cluster distances\cite{ros2023pdbi}.

\subsubsection*{Calinski-Harabas score (CH)} This is defined as the average similarity measure of each cluster with its most similar cluster, where similarity is the ratio of within-cluster distances to between-cluster distances\cite{wang2019improved}.


\subsubsection*{Average inter-cluster distance (I-CD)} This is the minimum distance between any two data points belonging to different clusters. The intercluster distance between clusters \(C_i\) and \(C_j\) using Euclidean distance can be expressed as:
\(d(C_i, C_j) = \sqrt{\sum_{k=1}^{n}(x_{ik} - x_{jk})^2}\), where \(x_{ik}\) and \(x_{jk}\) are data points in clusters \(C_i\) and \(C_j\), and \(n\) is the number of dimensions.

\subsubsection*{Average Variance} The generated cluster compactness and homogeneity can be derived through a measure of the variance of the cluster. The distribution of the time series over various clusters is observed to minimize the intracluster variance. For a cluster \(C_i\) belonging to a set of our final clustering, we compute its variance as shown below:

\(\text{ClusterVariance}(C_i) = \frac{1}{|C_i|} \sum_{x \in C_i} \lVert x - \mu_i \rVert^2\)
, where \(C_i\) is the cluster, \(|C_i|\) is the number of data points in cluster \(C_i\), \(x\) represents a data point in the cluster, and \(\mu_i\) is the centroid (mean) of cluster \(C_i\).

\subsubsection*{Average root mean squared error (RMSE)} For each observation, the error rate is the distance between every observation and its associated cluster centroid. The average total error is then the root sum of individual errors associated with each data point. \(\text{Average RMSE} = \frac{1}{k} \sum_{i=1}^{k} \sqrt{\frac{1}{n_i}\sum_{j=1}^{n_i}(d(x_j, \mu_i))^2}\)
\noindent , where
\(k\) is the number of clusters in the clustering solution,
\(n_i\) is the number of data points in cluster \(C_i\),
\(d(x_j, \mu_i)\) represents the distance between data point \(x_j\) in cluster \(C_i\) and the centroid \(\mu_i\) of that cluster.

\subsection{Stability Related Evaluation Metrics} \label{sec:Stability}
Cluster stability refers to the consistency of clustering results when the algorithm is run multiple times (e.g., with different initializations, subsamples of data, or different hyperparameters). A stable clustering algorithm will produce similar clusters across multiple runs.

\subsubsection*{Optimal Transport Alignment (OTA) for cluster stability}
Li et al.\cite{li2019optimal} recently developed an optimal transport framework for cluster stability. The approach seek ways to quantify the cost of transporting a probability distribution between two clusterings. 
They defined the stability between two clustering assignments \( C_1 \) and \( C_2 \) with \( n \) clusters each as the OT transport cost \( T \). \( T = \min_{\pi \in \Pi(C_1, C_2)} \sum_{i=1}^{n} \sum_{j=1}^{n} \pi_{ij} \, d(c_i^{(1)}, c_j^{(2)})\)
, where \( \pi \) is the transport plan, which specifies how much mass (or points) from cluster \( i \) in \( C_1 \) is transported to cluster \( j \) in \( C_2 \). \( d(c_i^{(1)}, c_j^{(2)}) \) is the cost (e.g., Euclidean distance) of transporting between clusters \( c_i^{(1)} \) in \( C_1 \) and \( c_j^{(2)} \) in \( C_2 \).
\( \Pi(C_1, C_2) \) represents the set of all valid transport plans between clusters in \( C_1 \) and \( C_2 \).
\subsubsection*{Figure of Merit (FoM) for Clustering Stability}
The FoM quantifies the stability of the clustering algorithm by measuring the average absolute difference between the pairwise distance matrices of the base and perturbed clustering results \cite{ben2001stability}. A lower FoM indicates higher stability. Given base clustering result \( P_{\text{base}} \) and perturbed clustering result \( P_{\text{perturbed}} \), the Figure of Merit (FoM) is defined as: \(
\text{FoM} = \frac{1}{n^2} \sum_{i=1}^{n} \sum_{j=1}^{n} \left| D_{\text{base}}(i, j) - D_{\text{perturbed}}(i, j) \right|\)
, where \( n \) is the number of data points, \( D_{\text{base}}(i, j) \) is the distance between points \( i \) and \( j \) in the base clustering \( P_{\text{base}} \), \( D_{\text{perturbed}}(i, j) \) is the distance between points \( i \) and \( j \) in the perturbed clustering \( P_{\text{perturbed}} \).


\subsubsection*{Average Proportion of Non-overlap (APN)}
The APN measures the proportion of data points that are assigned to different clusters in the two cluster assignments \( C_1 \) and \( C_2 \). A higher APN value indicates less stability, as more data points are assigned to different clusters. Given two cluster assignments \( C_1 \) and \( C_2 \) of a dataset with \( n \) data points, the Average Proportion of Non-overlap (APN) is defined as:
\(
\text{APN}(C_1, C_2) = \frac{1}{n} \sum_{i=1}^{n} \mathbb{I}(C_1(i) \neq C_2(i))
\)
, where: \( n \) is the total number of data points, \( C_1(i) \) and \( C_2(i) \) are the cluster labels assigned to the \( i \)-th data point by the cluster assignments \( C_1 \) and \( C_2 \), respectively and \( \mathbb{I}(C_1(i) \neq C_2(i)) \) is an indicator function that equals 1 if \( C_1(i) \neq C_2(i) \) and 0 otherwise.

\section{Experiments} \label{sec:Experiments}
All models are executed on AWS cloud environment using 5GB of S3 storage with 30 GB of ml.g4dn.xlarge GPU. The hardware used is a macOS ventura version 13.3, 16 GB, M1 pro chip. We applied the same python library across all models for homogeneity.
\subsection{Dataset and Data Preprocessing} \label{preprocesss}
\begin{table}[H]
\caption{CARRA - Data Description.}
\begin{center}
\label{tab:my-table3}
\setlength\tabcolsep{3pt}
\begin{tabular}{|l|l|l|l|}
\hline
\textbf{Var} & \textbf{Variable}         & \textbf{Range}                    & Unit             \\ \hline
tp           & Total precipitation       & {[}0, 0.0014{]}            & \(m\)    \\ \hline
rsn          & Snow density              & {[}99.9, 439.9{]}         & \(kg/m^3\)  \\ \hline
strd & Surface long-wave& {[}352599.1, 1232191.0{]} & \(J/m^2\) \\ \hline
t2m          & 2m  temperature           & {[}224.5, 289.4{]}        & \(k\)    \\ \hline
smlt         & Snowmelt                  & {[}-2.9\(e^{11}\),  8.5\(e^{04}\) {]} & \(m\)       \\ \hline
skt          & Skin temperature          & {[}216.6, 293.3{]}        & k \\ \hline
u10          & 10m u-wind & {[}-9.4, 13.3{]}        & \(m/s\)   \\ \hline
v10          & 10m v-wind& {[}-22.4, 16.1{]}       & \(m/s\)  \\ \hline
tcc          & Total cloud cover         & {[}0.0, 1.0{]} & \((0 - 1)\) \\ \hline
sd           & Snow depth                & {[}0, 6.5{]} & \(m\)      \\ \hline
msl          & Mean sea level pres   & {[}97282.1, 105330.8{]}  & \(Pa\)     \\ \hline
ssrd & Surface short-wave & {[}0, 1670912.0{]}      & \(J/m^2\) \\ \hline
\end{tabular}
\end{center}
\end{table}
To ensure generalizability, we experimented with three distinct multivariate spatiotemporal datasets. These are: The C3S Arctic Regional Reanalysis (CARRA) dataset contains 3-hourly analyses and hourly short term forecasts of atmospheric and surface meteorological variables at 2.5 km resolution~\cite{CARRA}, European Centre for Medium-Range Weather Forecasts (ECMRWF) ERA-5 global reanalysis product~\cite{ERA5}, daily atmospheric observations interpolated to pressure surfaces during the entire year of 2019, and span a spatial coverage of 2.5 degree x 2.5 degree global grids [\(144, 73\)]~\cite{NCAR}. For space limitation, we present only the CARRA data. All datasets follow the same preprocessing steps. 
Table \ref{tab:my-table3} presents a list of variables and their ranges selected for their impact on snow melt. The data consists of daily observations over the course of one year and is represented in four dimensions: longitude, latitude, time, and variables, with dimensions of [\(8, 18, 365, 13\)] respectively. When we directly convert the data into a 2D tabular data frame, the total feature count for each record would be 1872( \(8 \times 18 \times 13\) ), which from observation is high dimensional. After further exploring the dataset, we found the presence of null values which may occur from sensor malfunction or any physical conditions. The null values are replaced by the overall mean of the dataset so as not to obstruct the temporal pattern, which obviously will change the actual behavior of variable in the dataset. From Table \ref{tab:my-table3} the value ranges of the variables are
different. It is necessary to have the features fall within the same range better feature learning. To achieve this, we apply standard Min-Max Normalization (MMN) normalization
to rescale all features to fall within the range of [0 to 1].

\textbf{Other Models.} The details for our \textit{base models} that make up the building block for our homogeneous are shared on GitHub. To evaluate the performance of our proposed model, we compare across existing state-of-the-art ensemble models. These include spectral ensemble clustering~\cite{liu2015spectral}, parea~\cite{PFEIFER2023104406} and cluster ensemble~\cite{ghosh2011cluster}.
These models are selected partly because their approaches closely match our models' approach, considering the ensemble nature, and partly because their implementations are readily available online. We finetuned their hyperparameters to fit our data.
\subsection{Experiment Result}\label{sec:results}
Our goal is to design an ensemble model that can effectively capture the intrinsic and complex temporal patterns in high dimensional spatiotemporal data without access to the raw data or the models that generated the intermediary clusters. To evaluate the performance of our ensemble model, we conducted experiments on three real-world multivariate spatiotemporal datasets. Additionally, we tested the model’s robustness by performing stability experiments across multiple runs with different initializations and subsampling strategies. Our results show that the ensemble approach successfully captures intricate temporal dependencies in complex datasets, outperforming baseline methods on both performance and stability metrics.

\subsubsection{Performance-based results}
Given the absence of ground truth labels, we assessed the ability of the model to identify meaningful temporal clusters by using internal validation metrics. These metrics helped evaluate the model's cluster compactness and separation. Table \ref{tab:my-finalP} depicts the experimental results.
\begin{table}[H]
\begin{center}
\caption{Performance Evaluation of our proposed model: We compare HEDGTC against three baseline ensemble models on three spatiotemporal dataset using six internal cluster evaluation metrics. Far right are the results of HEDGTC.}
\label{tab:my-finalP}
\resizebox{\columnwidth}{!}{%
\begin{tabular}{|cccccc|}
\hline
\multicolumn{1}{|c|}{} &  \multicolumn{4}{c|}{Baseline Ensemble Models} &  Ours \\ \hline
\multicolumn{1}{|c|}{Data} &  \multicolumn{1}{c|}{Performance} &  \multicolumn{1}{c|}{ESC} &  \multicolumn{1}{c|}{Parea} &  \multicolumn{1}{c|}{\begin{tabular}[c]{@{}c@{}}Cluster\\ Ensemble\end{tabular}} &  HEDGTC \\ \hline
\multicolumn{1}{|c|}{\multirow{6}{*}{\begin{tabular}[c]{@{}c@{}}ERA5\\ \\ \\ \\ 7 optimal\\ clusters\end{tabular}}} &  \multicolumn{1}{c|}{Silhouette $\uparrow$} &  \multicolumn{1}{c|}{0.2337} &  \multicolumn{1}{c|}{0.2318} &  \multicolumn{1}{c|}{0.2246} &  \textbf{0.3773} \\ \cline{2-6} 
\multicolumn{1}{|c|}{} &  \multicolumn{1}{c|}{DB $\downarrow$} &  \multicolumn{1}{c|}{1.7731} &  \multicolumn{1}{c|}{1.7532} &  \multicolumn{1}{c|}{1.6722} &  \textbf{1.3766} \\ \cline{2-6} 
\multicolumn{1}{|c|}{} &  \multicolumn{1}{c|}{CH $\uparrow$} &  \multicolumn{1}{c|}{88.9491} &  \multicolumn{1}{c|}{98.4634} &  \multicolumn{1}{c|}{79.3687} &  \textbf{98.6553} \\ \cline{2-6} 
\multicolumn{1}{|c|}{} &  \multicolumn{1}{c|}{RMSE $\downarrow$} &  \multicolumn{1}{c|}{14.2133} &  \multicolumn{1}{c|}{13.7791} &  \multicolumn{1}{c|}{\textbf{10.4307}} &  13.7708 \\ \cline{2-6} 
\multicolumn{1}{|c|}{} &  \multicolumn{1}{c|}{Var $\downarrow$} &  \multicolumn{1}{c|}{0.1039} &  \multicolumn{1}{c|}{0.1030} &  \multicolumn{1}{c|}{\textbf{0.0323}} &  0.1030 \\ \cline{2-6} 
\multicolumn{1}{|c|}{} &  \multicolumn{1}{c|}{I-CD $\uparrow$} &  \multicolumn{1}{c|}{5.3526} &  \multicolumn{1}{c|}{6.5680} &  \multicolumn{1}{c|}{4.2369} &  \textbf{6.8952} \\ \hline
\multicolumn{6}{|c|}{} \\ \hline
\multicolumn{1}{|c|}{\multirow{6}{*}{\begin{tabular}[c]{@{}c@{}}CARRA\\ \\ \\ 5 optimal\\ clusters\end{tabular}}} &  \multicolumn{1}{c|}{Silhouette $\uparrow$} &  \multicolumn{1}{c|}{0.2159} &  \multicolumn{1}{c|}{0.1883} &  \multicolumn{1}{c|}{0.1967} &  \textbf{0.2820} \\ \cline{2-6} 
\multicolumn{1}{|c|}{} &  \multicolumn{1}{c|}{DB $\downarrow$} &  \multicolumn{1}{c|}{1.7803} &  \multicolumn{1}{c|}{1.5617} &  \multicolumn{1}{c|}{1.6520} &  \textbf{1.5206} \\ \cline{2-6} 
\multicolumn{1}{|c|}{} &  \multicolumn{1}{c|}{CH $\uparrow$} &  \multicolumn{1}{c|}{65.5666} &  \multicolumn{1}{c|}{64.8821} &  \multicolumn{1}{c|}{78.3657} &  \textbf{78.4460} \\ \cline{2-6} 
\multicolumn{1}{|c|}{} &  \multicolumn{1}{c|}{RMSE $\downarrow$} &  \multicolumn{1}{c|}{5.5571} &  \multicolumn{1}{c|}{\textbf{5.5723}} &  \multicolumn{1}{c|}{5.8827} &  5.5787 \\ \cline{2-6} 
\multicolumn{1}{|c|}{} &  \multicolumn{1}{c|}{Var $\downarrow$} &  \multicolumn{1}{c|}{\textbf{0.0160}} &  \multicolumn{1}{c|}{\textbf{0.0160}} &  \multicolumn{1}{c|}{\textbf{0.0160}} &  \textbf{0.0160} \\ \cline{2-6} 
\multicolumn{1}{|c|}{} &  \multicolumn{1}{c|}{I-CD $\uparrow$} &  \multicolumn{1}{c|}{3.2677} &  \multicolumn{1}{c|}{3.2410} &  \multicolumn{1}{c|}{2.8043} &  \textbf{3.3919} \\ \hline
\multicolumn{6}{|c|}{} \\ \hline
\multicolumn{1}{|c|}{\multirow{6}{*}{\begin{tabular}[c]{@{}c@{}}NCAR\\ Reanalysis 1\\ \\ 7 optimal\\ clusters\end{tabular}}} &  \multicolumn{1}{c|}{Silhouette $\uparrow$} &  \multicolumn{1}{c|}{0.3906} &  \multicolumn{1}{c|}{0.4228} &  \multicolumn{1}{c|}{0.5613} &  \textbf{0.6149} \\ \cline{2-6} 
\multicolumn{1}{|c|}{} &  \multicolumn{1}{c|}{DB $\downarrow$} &  \multicolumn{1}{c|}{0.9869} &  \multicolumn{1}{c|}{0.9568} &  \multicolumn{1}{c|}{0.7655} &  \textbf{0.7275} \\ \cline{2-6} 
\multicolumn{1}{|c|}{} &  \multicolumn{1}{c|}{CH $\uparrow$} &  \multicolumn{1}{c|}{503.2038} &  \multicolumn{1}{c|}{549.0724} &  \multicolumn{1}{c|}{851.8517} &  \textbf{852.974} \\ \cline{2-6} 
\multicolumn{1}{|c|}{} &  \multicolumn{1}{c|}{RMSE $\downarrow$} &  \multicolumn{1}{c|}{4.0202} &  \multicolumn{1}{c|}{3.8657} &  \multicolumn{1}{c|}{3.1591} &  \textbf{3.0297} \\ \cline{2-6} 
\multicolumn{1}{|c|}{} &  \multicolumn{1}{c|}{Var $\downarrow$} &  \multicolumn{1}{c|}{\textbf{0.1770}} &  \multicolumn{1}{c|}{\textbf{0.1770}} &  \multicolumn{1}{c|}{\textbf{0.1770}} &  \textbf{0.1770} \\ \cline{2-6} 
\multicolumn{1}{|c|}{} &  \multicolumn{1}{c|}{I-CD $\uparrow$} &  \multicolumn{1}{c|}{0.8450} &  \multicolumn{1}{c|}{0.8720} &  \multicolumn{1}{c|}{0.8585} &  \textbf{0.8946} \\ \hline
\end{tabular}%
}
\end{center}
\end{table}

\subsubsection{Stability-based results}
\begin{figure}[H]
  \centering
  \includegraphics[width=0.45\textwidth]{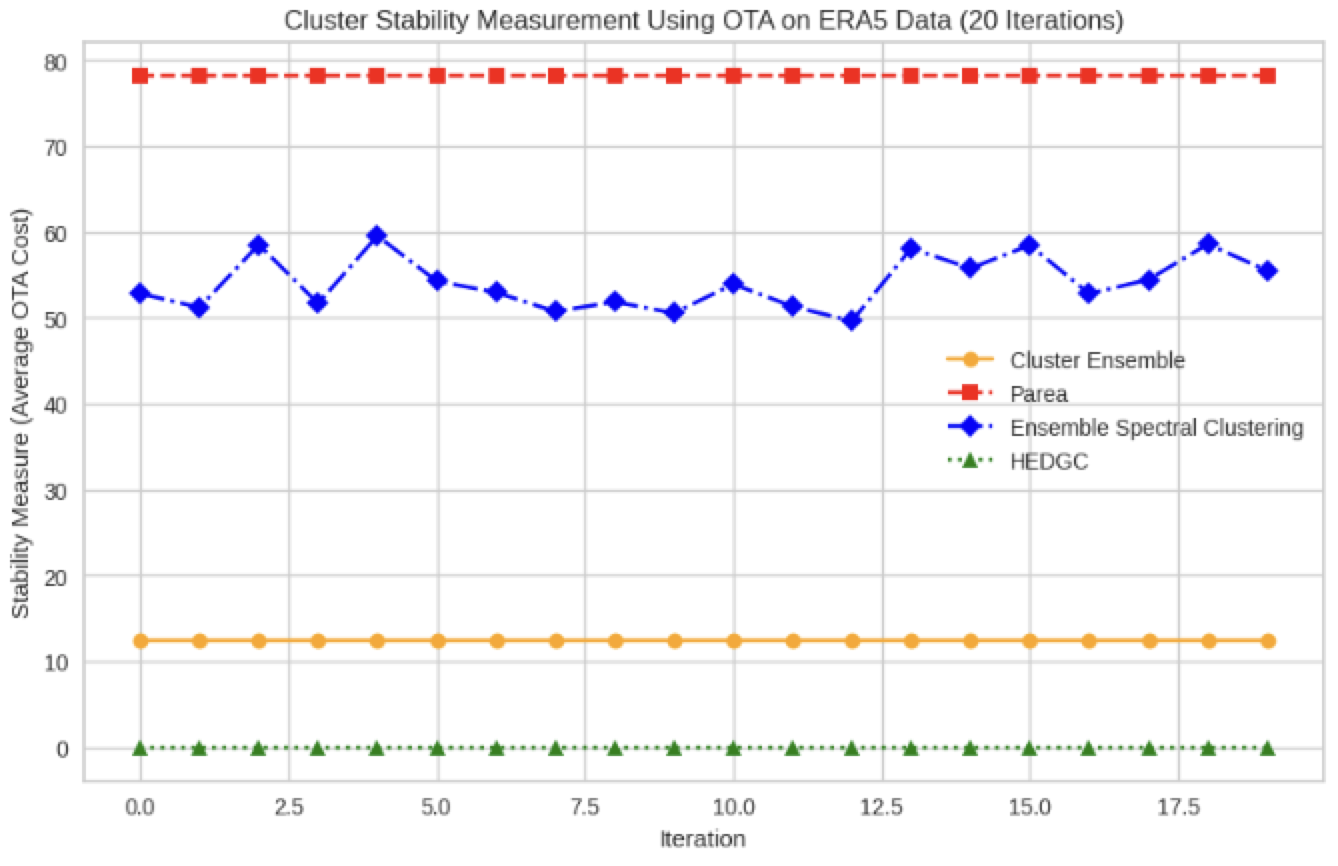}
  \centering \caption{OTA Stability Assessment on ERA5 on 20 executions. HEDGTC (green line) is seen to outperform all other baseline ensemble models with the lowest OTA. The higher the OTA measurement score, the less stable is the algorithm. The Ensemble Spectral Clustering model displays a wave-like stability structure as a results of the initial seed problem, noise introduced during perturbation and dependence on graph construction. Parea is least stable partly due to the initial seeds problem and the absence of consensus matrix post-processing. }
  \label{fig:stable}
\end{figure}

Table \ref{tab:Stability} presents a numerical evaluation of how stable our proposed model is, when compared to baseline models across all three datasets. HEDGTC outperforms other baseline models in all three stability measures across all three datasets. Both HEDCC and baselines used in this experiment were executed 20 times and the average considered. For space reasons, we present only the visualization of the stability results from \(OTA\) on all ensemble models when executed 20 times on ERA5 data as seen in Figure \ref{fig:stable}.
\begin{table}[H]
\begin{center}
\caption{\textbf{Stability Evaluation of our proposed model:} HEDGTC produced more stable results across three stability measures on three spatiotemporal datasets when compared to existing state of the art ensemble models.}
\label{tab:Stability}
\resizebox{\columnwidth}{!}{%
\begin{tabular}{lcclclclcl|}
\cline{3-10}
\multicolumn{2}{l|}{} &
  \multicolumn{6}{c|}{\textbf{Baseline Ensemble Models}} &
  \multicolumn{2}{c|}{\textbf{\begin{tabular}[c]{@{}c@{}}Proposed \\ Ensemble Model\end{tabular}}} \\ \hline
\multicolumn{1}{|c|}{Datasets} &
  \multicolumn{1}{c|}{\begin{tabular}[c]{@{}c@{}}Measure Used\\ for Stability\end{tabular}} &
  \multicolumn{2}{c|}{ESC} &
  \multicolumn{2}{c|}{Parea} &
  \multicolumn{2}{c|}{Cluster Ens} &
  \multicolumn{2}{c|}{HEDGTC} \\ \hline
\multicolumn{1}{|l|}{\multirow{3}{*}{\begin{tabular}[c]{@{}l@{}}ERA5 \\ 7 Optimal \\ Clusters\end{tabular}}} &
  \multicolumn{1}{c|}{OTA \( \downarrow \)} &
  \multicolumn{2}{c|}{55.86} &
  \multicolumn{2}{c|}{78.28} &
  \multicolumn{2}{c|}{12.41} &
  \multicolumn{2}{c|}{\textbf{0.00}} \\ \cline{2-10} 
\multicolumn{1}{|l|}{} &
  \multicolumn{1}{c|}{FOM \( \downarrow \)} &
  \multicolumn{2}{c|}{0.27} &
  \multicolumn{2}{c|}{0.28} &
  \multicolumn{2}{c|}{0.30} &
  \multicolumn{2}{c|}{\textbf{0.11}} \\ \cline{2-10} 
\multicolumn{1}{|l|}{} &
  \multicolumn{1}{c|}{APN \( \downarrow \)} &
  \multicolumn{2}{c|}{0.85} &
  \multicolumn{2}{c|}{0.84} &
  \multicolumn{2}{c|}{0.70} &
  \multicolumn{2}{c|}{\textbf{0.60}} \\ \hline
\multicolumn{10}{|l|}{} \\ \hline
\multicolumn{1}{|l|}{\multirow{3}{*}{\begin{tabular}[c]{@{}l@{}}CARRA\\ 5 optimal\\ clusters\end{tabular}}} &
  \multicolumn{1}{c|}{OTA \( \downarrow \)} &
  \multicolumn{2}{c|}{93.20} &
  \multicolumn{2}{c|}{105.50} &
  \multicolumn{2}{c|}{82.00} &
  \multicolumn{2}{c|}{\textbf{47.2}} \\ \cline{2-10} 
\multicolumn{1}{|l|}{} &
  \multicolumn{1}{c|}{FOM \( \downarrow \)} &
  \multicolumn{2}{c|}{0.32} &
  \multicolumn{2}{c|}{0.30} &
  \multicolumn{2}{c|}{0.29} &
  \multicolumn{2}{c|}{\textbf{0.25}} \\ \cline{2-10} 
\multicolumn{1}{|l|}{} &
  \multicolumn{1}{c|}{APN \( \downarrow \)} &
  \multicolumn{2}{c|}{0.84} &
  \multicolumn{2}{c|}{0.85} &
  \multicolumn{2}{c|}{0.89} &
  \multicolumn{2}{c|}{\textbf{0.76}} \\ \hline
\multicolumn{10}{|l|}{} \\ \hline
\multicolumn{1}{|l|}{\multirow{3}{*}{\begin{tabular}[c]{@{}c@{}}NCAR\\ 7 optimal\\ clusters\end{tabular}}} &
  \multicolumn{1}{c|}{OTA \( \downarrow \)} &
  \multicolumn{2}{c|}{93.20} &
  \multicolumn{2}{c|}{105.50} &
  \multicolumn{2}{c|}{82.00} &
  \multicolumn{2}{c|}{\textbf{17.04}} \\ \cline{2-10} 
\multicolumn{1}{|l|}{} &
  \multicolumn{1}{c|}{FOM \( \downarrow \)} &
  \multicolumn{2}{c|}{0.32} &
  \multicolumn{2}{c|}{0.30} &
  \multicolumn{2}{c|}{\textbf{0.28}} &
  \multicolumn{2}{c|}{0.35} \\ \cline{2-10} 
\multicolumn{1}{|l|}{} &
  \multicolumn{1}{c|}{APN \( \downarrow \)} &
  \multicolumn{2}{c|}{0.84} &
  \multicolumn{2}{c|}{0.85} &
  \multicolumn{2}{c|}{0.89} &
  \multicolumn{2}{c|}{\textbf{0.77}} \\ \hline
\end{tabular}%
}
\end{center}
\end{table}

\subsection{Discussions on Performance and Stability}\label{sec:discussions} \textbf{Performance.} Table \ref{tab:my-finalP} presents the final performance results based on selected internal cluster validation measures when applied to all three datasets. On all datasets, HEDGTC outperformed all baseline ensemble models as reported by Silhouette, DB, CH and I-CH. This implies HEDGTC was able to capture the underlying complex patterns in all three datasets with significant improvements on performance. ESC follows a homogeneous approach hence is bounded by the limitation of spectral clustering. Some include applying local information to globally cluster eigenvectors, disregard within cluster information, inability to cluster datasets of different structures and sizes by just using the first few eigenvectors of the generated adjacency matrices. These base clustering limitation extend to the limitations of ESC leading to sub-optimal clusterings. In addition, ESC achieves consensus through majority voting without no further matrix post processing which can further lead to sub-optimal results. In a similar way, Cluster Ensemble applies KMeans in a homogeneous fashion and inherits all shortcomings of centroid-based clustering algorithms. KMeans struggles with the problem of initial seeds and optimal number of clusters. Although the ensemble of KMeans algorithms seek to solve the problem of initialization, it does not guarantee optimal results. All datasets used in this experiment largely depend on weather and their inherent temporal fluctuations and hence subject to temporal clustering boundary delineation, a process crucial for understanding transitions between different climate states, such as shifts in temperature, precipitation patterns, or the onset and cessation of seasonal events. HEDGTC is able to capture these temporal clustering boundary delineation patterns. HEDGTC operates in a hybrid fashion by harnessing the power of both homogeneous and heterogeneous ensemble to produce optimal clusters. This approach mitigates the limitation seen in ESC, SC and Parea. To further improve performance, HEDGTC implements post-processing by deleting weak nodes, matrix normalization and addressing errors from misclassification from previous clusterings. On the other hand Cluster ensemble performed slightly better than HEDGTC on RMSE because it measures similarity by computing the distance between the centroids and not between data points hence its not affected by border points. The variance is constant across all models when tested on CARRA and NCAR Reanalysis data with slight improvement seen by Cluster Ensemble model on ERA5 data.



\textbf{Stability.} Table \ref{tab:Stability} depicts the average stability results after 20 iterations from OTA, FOM and APN when applied to all three datasets. All baseline ensemble models show high degree of instability. While Parea and Cluster Ensemble may  benefit from reducing the sensitivity to outliers, they both struggle with the initial seeds problem. Although we predefined the total number of clusters across the entire experiment, these algorithms are sensitive to initial position of the seeds which may introduce convergence issues increasing the risk to settle on a local minimum or not converge at all. Both algorithms are not robust against noise and stability can be compromised in the presence of noisy data. Both algorithms assume certain shapes of clusters which, in most cases are not present in high dimensional data. This assumption leads to unstable clusterings. While struggling with the above mentioned points, Ensemble Spectral Clustering depends on the construction of a similar graph and small changes in graph construction parameters can lead to significant changes in the clustering results, making the final ensemble result unstable. HEDGTC is more stable as shown by OTA, FOM and APN on Table \ref{tab:Stability} and OTA in Figure \ref{fig:stable}. The resulting merged adjacency matrix is fed into a three-layered stacked GAT autoencoder which extracts the node features, edge index and the weights of the neighboring nodes. Stability is further enhanced by projecting this information onto a lower dimension iteratively where clustering is performed. HEDGTC when applied to all three datasets showed improved stability as reported by OTA, FOM and APN stability measures.


\subsection{Computational complexity} 
\noindent The overall complexity is the product of invidual complexities. Let \(T_1\): Homogeneous, \(T_2\): Heterogeneous, \(T_3\): Final Clustering. For \(T_1\), time complexity largely depends on both the number of algorithm and the number of ensemble members
\(m\), hence 
\(O(m \cdot T(n))\). For \(T_2\), 
let's assume \(A_i\) is the \(i{th}\) clustering algorithm with time complexity \(T_i{(n)}\), the total time complexity for our heterogeneous ensemble clustering is: \(O(\sum_{i=1}^{m} T(n) + m \cdot n^2)\). For \(T_3\), the time complexity for co-occurrence consensus ensemble clustering is: \(O(\sum_{i=1}^{m} T(i) + m \cdot n^2 + n^3)\) and overall time complexity for NMF consensus ensemble clustering is: \(O(m \cdot n \cdot r\cdot t\cdot m\cdot n^2\)) where \(r =\) rank and \(t = \) number of iterations. Hence the time complexity at phase \(T_3\) is \(O(\sum_{i=1}^{m} T(i) + m\cdot n^2 + n^3)\) + \(O(m \cdot n \cdot r\cdot t\cdot m\cdot n^2\)) resulting to the dominant term of \(O(n^3)\).

\subsection{Ablation study} 
Table \ref{tab:my-table8} presents performance-based and stability-based results of a quantitative comparative study achieved by systematically isolating and examining the effects of individual components of HEDGTC experimented on ERA5 data.

\begin{table}[H]
\caption{Ablation Study: We show the importance of various subcomponents of HEDGTC using both Performance and Stability measures when applied on ERA5 Data.}
\begin{center}
\label{tab:my-table8}
\resizebox{\columnwidth}{!}{%
\begin{tabular}{|cccclll}
\hline
\multicolumn{7}{|c|}{Performance - based} \\ \hline
\multicolumn{1}{|c|}{\textbf{}} &
  \multicolumn{1}{c|}{\textbf{Silh} $\uparrow$} &
  \multicolumn{1}{c|}{\textbf{DB} $\uparrow$} &
  \multicolumn{1}{c|}{\textbf{CH} $\uparrow$} &
  \multicolumn{1}{c|}{\textbf{RMSE} $\downarrow$} &
  \multicolumn{1}{c|}{\textbf{Var} $\downarrow$} &
  \multicolumn{1}{c|}{\textbf{I-CD} $\uparrow$} \\ \hline
\multicolumn{1}{|c|}{HEDGTC$_{nmf}$} &
  \multicolumn{1}{c|}{0.3532} &
  \multicolumn{1}{c|}{1.4549} &
  \multicolumn{1}{c|}{101.5804} &
  \multicolumn{1}{c|}{\textbf{13.6453}} &
  \multicolumn{1}{c|}{0.1031} &
  \multicolumn{1}{c|}{6.4173} \\ \hline
\multicolumn{1}{|c|}{HEDGTC$_{co-occ}$} &
  \multicolumn{1}{c|}{0.3639} &
  \multicolumn{1}{c|}{1.4207} &
  \multicolumn{1}{c|}{\textbf{93.1555}} &
  \multicolumn{1}{c|}{14.0164} &
  \multicolumn{1}{c|}{0.1031} &
  \multicolumn{1}{c|}{\textbf{7.4525}} \\ \hline
\multicolumn{1}{|c|}{HEDGTC} &
  \multicolumn{1}{c|}{\textbf{0.3773}} &
  \multicolumn{1}{c|}{\textbf{1.3766}} &
  \multicolumn{1}{c|}{98.6553} &
  \multicolumn{1}{c|}{13.7708} &
  \multicolumn{1}{c|}{\textbf{0.1030}} &
  \multicolumn{1}{c|}{6.8952} \\ \hline
\multicolumn{4}{|l|}{} &
   &
   &
   \\ \cline{1-4}
\multicolumn{4}{|c|}{Stability - based} &
   &
   &
   \\ \cline{1-4}
\multicolumn{1}{|c|}{} &
  \multicolumn{1}{c|}{\textbf{OTA} $\downarrow$} &
  \multicolumn{1}{c|}{\textbf{FOM} $\downarrow$} &
  \multicolumn{1}{c|}{\textbf{APN} $\downarrow$} &
   &
   &
   \\ \cline{1-4}
\multicolumn{1}{|c|}{HEDGTC$_{nmf}$} &
  \multicolumn{1}{c|}{\textbf{17.04}} &
  \multicolumn{1}{c|}{0.35} &
  \multicolumn{1}{c|}{0.77} &
   &
   &
   \\ \cline{1-4}
\multicolumn{1}{|c|}{HEDGTC$_{co-occ}$} &
  \multicolumn{1}{c|}{22.00} &
  \multicolumn{1}{c|}{0.14} &
  \multicolumn{1}{c|}{\textbf{0.57}} &
   &
   &
   \\ \cline{1-4}
\multicolumn{1}{|c|}{HEDGTC} &
  \multicolumn{1}{c|}{\textbf{0.00}} &
  \multicolumn{1}{c|}{0.11} &
  \multicolumn{1}{c|}{0.60} &
   &
   &
   \\ \cline{1-4}
\end{tabular}%
}
\end{center}
\end{table}

In Table \ref{tab:my-table8}, \(HEDGTC_{nmf}\) is a variant without consensus through Object Co-occurrence and \({HEDGTC_{co-occ}}\) is a variant without consensus through Non-negative matrix factorization. The ablation study results proves that integrating both consensus functions improves the ratio between the cohesion and separation depicted by the significant increase of the silhouette score and davis bouldin score. Although there is a sharp decrease in the average distance between data points and their respective centroids as captured by the RMSE and the inter-cluster distance, the overall cluster variance proved stable with slight improvements. From our stability measurement, while the OTA and APN respectively prove individual component stability, FOM and OTA prove that an ensemble of both components improved the our model's stability. All reported numbers are averages after 20 iterations. \(HEDGTC_{nmf}\) benefits from dimensionality reduction while \({HEDGTC_{co-occ}}\) benefits from noise reduction and matrix normalization, both post matrix processing seek to improve stability.


\section{Conclusion}\label{sec:Conclusion}
In this paper, we proposed an end-to-end Hybrid Ensemble Deep Graph Temporal Clustering (HEDGTC) model to cluster complex multivariate spatiotemporal data without accessing the features or algorithms that determined the base partitions. HEDGTC is able to learn decision boundaries and generate good quality partitions.
To improve performance and stability, HEDGTC integrates a list of base clustering models to learn decision boundaries characterized by intrinsic features and semantic patterns of complex multidimensional spatiotemporal data. It further adopts and implements a dual consensus approach that merge resulting matrices into a unified matrix suitable for graph clustering. The final matrix is fed into a graph attention autoencoder and KMeans is applied to the dense layer to yield our final clusterings.

\bibliographystyle{acm}
\bibliography{biblo}
\end{document}